\documentclass[twoside,11pt]{article}

\usepackage{jmlr2e}
\usepackage{hyperref}       
\usepackage{url}            
\usepackage{booktabs}       
\usepackage{amsfonts}       
\usepackage{nicefrac}       
\usepackage{microtype}      
\usepackage{graphicx}
\usepackage{natbib}
\PassOptionsToPackage{numbers, compress}{natbib}
\usepackage{amsmath}
\usepackage{mathtools}

\usepackage{color}
\usepackage{subcaption}
\usepackage{multirow}
\usepackage[ruled,vlined]{algorithm2e}

\usepackage{lastpage}
\jmlrheading{20}{2019}{1-\pageref{LastPage}}{6/18; Revised
	4/19}{11/19}{18-374}{Michael Minyi Zhang and Sinead Williamson}
\ShortHeadings{Embarrassingly Parallel Inference for Gaussian Processes}{Zhang and Williamson}

\firstpageno{1}

\begin{document}

\title{Embarrassingly Parallel Inference for Gaussian Processes}

\author{\name Michael Minyi Zhang \email mz8@cs.princeton.edu \\
       \addr Department of Computer Science\\
		Princeton University\\
		Princeton, NJ 08544, USA
       \AND
       \name Sinead A. Williamson \email sinead.williamson@mccombs.utexas.edu\\
	\addr Department of Statistics and Data Science\\
	Department of Information, Risk and Operations Management\\
	The University of Texas at Austin\\
	Austin, TX 78712, USA}

\editor{Manfred Opper}

\maketitle

\begin{abstract}
	Training Gaussian process-based models typically involves an $ O(N^3)$ computational bottleneck due to inverting the covariance matrix. Popular methods for overcoming this matrix inversion problem cannot adequately model all types of latent functions, and are often not parallelizable. However, judicious choice of model structure can ameliorate this problem. A mixture-of-experts model that uses a mixture of $K$ Gaussian processes offers modeling flexibility and opportunities for scalable inference. Our embarrassingly parallel algorithm combines low-dimensional matrix inversions with importance sampling to yield a flexible, scalable mixture-of-experts model that offers comparable performance to Gaussian process regression at a much lower computational cost.

\end{abstract}


\section{Introduction}\label{sec:intro}

Gaussian processes (GPs) provide a flexible family of distributions over functions, that have been widely adopted for problems including regression, classification and optimization due to their ease of use in modeling latent functions. Additional flexibility can be achieved using mixture-of-experts models \citep{Gramacy:Lee:2005,Rasmussen:Ghahramani:2002,Meeds:Osindero:2006}, which use a mixture of Gaussian processes. Different mixture components can have different covariance patterns, allowing for non-stationarity in the resulting function without resorting to explicitly non-stationary covariance functions.

Unfortunately, the flexibility of Gaussian processes and related models comes at a cost. Inference in GP models with $N$ observations involves  repeated inversion of an $N\times N$ matrix, which typically scales $O\left( N^3 \right)$. Mixture-of-experts models fare slightly better since, conditioned on the partition, we are inverting a block-diagonal matrix; however the computational savings are tempered by the cost of averaging over partitions, typically using MCMC. This computational bottleneck has thus far prevented Gaussian processes and mixtures-of-experts from being used in so-called ``big data'' situations.


Two main approaches have been proposed to ameliorate the computational complexity for inference in the simple Gaussian process regression setting: sparse methods, that aim to reduce the size of the matrix to be inverted, and local methods, that aim to simplify its structure. Unfortunately, both methods exhibit key failure modes as we reduce the computational cost: local methods can miss long-range correlations, and sparse methods tend to miss short-range fluctuations.  Further, methods of these types are not typically parallelizable to run efficiently on a distributed architecture.

In this paper, we propose a novel inference algorithm for fitting mixtures of partitioned Gaussian processes that is flexible and easily distributed. The ``Importance Sampled Mixture of Experts'' (IS-MOE) uses importance sampling to average over Gaussian processes with block-diagonal covariance matrices.  A location-based distribution over partitions allows us to capture non-stationarity. We learn, in parallel, multiple partitioned Gaussian processes sampled from the conditional distribution over partitions given the input locations thereby allowing us to take advantage of the lower inversion cost of a block-diagonal matrix. Minibatch-based stochastic approximations reduce the size of these blocks further, while maintaining competitive performance. Importance weights are also calculated in a distributed manner, with the only global communication occurring when the importance-weighted samples are combined.  The resulting posterior predictive distribution has a more expressive expected covariance matrix than a single block-diagonal matrix, avoiding edge effects common with local methods and allowing for an expressive covariance structure that can model both long- and short-range covariance as well as non-stationary behavior in the latent function. 

We start in Section~\ref{sec:background} by reviewing the Gaussian process and describing existing methods for scaling inference. We then describe our approach in Section~\ref{sec:method}, before presenting detailed experimental evaluation in Section~\ref{sec:results} to showcase its efficacy versus existing methods. Further extensions and applications are discussed in Section~\ref{sec:conclusion}.

\section{Background}\label{sec:background}
\subsection{Gaussian processes and related models}
A Gaussian process is a distribution over functions $f: \mathbb{R}^D\rightarrow \mathbb{R}$, parametrized by some mean function $m(x)$, typically taken as zero, and a covariance function $\Sigma(x,x')$. For a given $m$ and $\Sigma$, a GP is a unique distribution over functions $f$ such that for any finite set of points, $x_1,\dots, x_N\in \mathbb{R}^D$, the function evaluated at those points is multivariate normally distributed with mean and covariance given by $m$ and $\Sigma$ evaluated at these inputs.

This distribution over functions can be used in a variety of applications, including regression, classification and optimization \citep[see for example][]{Rasmussen:Williams:2006,Snoek:Larochelle:Adams:2012,Wang:Fleet:Hertzmann:2005}. For simplicity, we focus here on the regression setting, where a function $f\sim\mbox{GP}(0,\Sigma)$ maps our inputs $X=(x_i)_{i=1}^N$ to our outputs $Y=(y_i)_{i=1}^N$, such that $y_i \sim \mbox{N}(f(x_i), \sigma^2)$.

In this setting, the posterior distribution of $f$ given $X$, $Y$ and $\Sigma$ is analytically tractable, and the inference challenge reduces to inferring the hyperparameters, $\Theta$, that control the form of the covariance function. Optimizing or sampling these hyperparameters involves inverting the covariance matrix $\boldsymbol{\Sigma}$ obtained by evaluating $\Sigma(\cdot,\cdot)$ at the inputs $x_1,\dots, x_N$. In general, the computational cost of inverting this matrix is $O(N^3)$.

Mixture-of-experts models  \citep{Jacobs:1991} are a hierarchical extension of Gaussian processes, that model each output $y_i$ using a mixture of Gaussian processes. This mixture can be specified either by a distribution over partitions of the input space \citep{Rasmussen:Ghahramani:2002, Gramacy:Lee:2005, Yuan:Neubauer:2009} or on the joint space of inputs and outputs \citep{Meeds:Osindero:2006}. This offers two advantages over a single Gaussian process. First, each composite Gaussian process can have a separate covariance function, allowing us to capture different behaviors in different regions. Second, conditioned on the partition we have $K$ independent Gaussian processes with average size $N/K$, reducing the computational cost of matrix inversion.

Unfortunately, this computational advantage is counterbalanced by the  computational cost of inferring the distribution over partitions, which is done using either MCMC \citep{Rasmussen:Ghahramani:2002,Gramacy:Lee:2005,Meeds:Osindero:2006} or variational methods \citep{Yuan:Neubauer:2009}. As a result,  mixture of expert models are typically not considered ``scalable''. Performing MCMC-based inference over partitions can be expensive and while variational methods are generally faster, the Markovian relationship between samples precludes direct parallelization.

\subsection{Scalable inference methods for Gaussian processes}\label{sec:sparse_local}

Most scalable inference approaches focus on reducing the $O(N^3)$ cost of covariance matrix inversion. Two broad classes of methods have been proposed: ``sparse'' methods which parametrize the covariance based on $M<<N$ inducing inputs, and ``local'' methods that replace the dense $N\times N$ covariance matrix with a block-diagonal matrix. 

Sparse GP approximations parameterize the covariance matrix of the GP model with $M$ pseudo-inputs, where $M<<N$. The pseudo-input locations are chosen so that the posterior function evaluated at these points is a good approximation to the true posterior, for example by maximum likelihood optimization \citep{Snelson:Ghahramani:2005} or variational inference \citep{Titsias:2009}. The computational saving comes from replacing an $N\times N$ covariance matrix with an $M\times M$ matrix. In the regression case, this reduces the training cost to $O(NM^2)$. Further computational savings can be obtained by using stochastic variational inference (SVI) to update the inducing points by calculating necessary gradients based only on size-$B$ subsets of the $N$ datapoints, reducing computational cost to $O\left(M^2\min \left\{ M,B \right\}\right)$ \citep{Hensman:2013}. While sparse methods can yield impressive speed-ups, they tend to have a decreased ability to model high-frequency fluctuations in the function, since the number of inducing points limits the amount of variation we can capture. Additionally, as with the full-covariance GP it approximates, the sparse GP cannot naturally model non-stationary data without resorting to a non-stationary kernel.

Local Gaussian process methods make local approximations to the dense covariance matrix so that a low-rank representation of the covariance matrix is inverted instead of the full-rank matrix. Mixture-of-experts models, described above, fall under this framework, since conditioned on the partition we have a block-diagonal covariance matrix; however the cost of averaging over partitions means these are not generally seen as scalable models. Product-of-experts models \citep{Tresp:2000,Cao:Fleet:2014,Ng:Deisenroth:2014,Deisenroth:Ng:2015} avoid this by using a single partition, and avoid edge effects by multiplying the predictions of the local Gaussian processes. Conditioned on the partitioning, inference in the local GP scales approximately as $O(N^3/K^2)$, since we need to invert $K$ matrices of average size $\frac{N}{K} \times \frac{N}{K}$. \citet{Park:Huang:Ding:2011} also use a block-diagonal approximation, and use a boundary value function to ensure continuity between regions. 
 
Taking a different perspective, \cite{Kaufman:2008} applies a ``tapering'' function to the covariance matrix so that observation pairs with low correlation are set to zero and provides theorems for estimator consistency when the covariance function used is a Mat\'{e}rn kernel. \cite{Gramacy:Apley:2015} try to learn the local approximation by taking the $ n $-nearest neighbors of a predictive value $ X^{\ast} $ to the data $ X $ and learns both the function hyperparameters and predictive distribution jointly by iteratively increasing the size of the nearest neighbors until a stopping criteria is satisfied for all predictive inputs. 


As noted in \cite{Low:2015}, local methods will be good at capturing short-range correlations, where the correlation structure is well approximated. Further, if a block-diagonal covariance is used, they allow us to use  different covariance hyperparameters in different blocks, capturing behavior which is locally approximately stationary, but where the lengthscale varies across the input space  \citep{Tresp:2000,Rasmussen:Ghahramani:2002}. This is in contrast with sparse methods, where the number of inducing points limits the ability to learn very short-range correlations,  and which can only capture non-stationarity if we use an explicitly non-stationary covariance function.

The disadvantage of the local methods, however, is that they risk ignoring important correlations. For example, the block-diagonal approaches  assume zero correlation between different blocks in the partition. If the data points are partitioned based on location, this means that long-range correlations will be ignored; if they are partitioned randomly, the model will tend to perform poorly if the number of observations in some region of $\mathbb{R}^D$ is low. Moreover, \cite{Liu:Cai:Wang:Ong:2018} show that local methods like the RBCM will systematically produce overconfident predictions which violates one of the major benefits of using Bayesian methods in the first place--that being proper uncertainty quantification. 

\subsection{Distributed inference for Gaussian processes}\label{sec:dist_bg}

The sparse and local approximations described above aim to reduce the overall computational burden by reducing the size of matrices to be inverted. When run on a single machine, this reduction in computational cost leads directly to faster inference. However, we may also be interested in distributing computation cost across multiple threads or machines. Even if the total computational cost is the same, we can reduce total time by distributing computation onto multiple parallel threads. Alternatively, if we increase the computational budget then we may be able to improve our posterior estimate by running multiple samplers in parallel and then combining the results without increasing the time budget.  

Local partition-based GP methods that do not average over partitions, such as product-of-experts models are well suited to this sort of parallelism. They split a single GP problem into $K$ independent problems whose parameters can be inferred in parallel. We only need to communicate between the $K$ subproblems at the end when we combine their predictions. This type of algorithm, where global communication occurs only once after all the local computation is complete, is known as ``embarrassingly parallel''. \citet{Ng:Deisenroth:2014} exploit these independences, in a weighted product-of-experts model, to obtain a distributable algorithm appropriate for large datasets. 

\subsection{Fast Bayesian inference via stochastic approximations}\label{sec:minibatch}

When performing Bayesian inference on large datasets, much of the computational cost is due to calculating functions of the data -- for example, gradients or likelihoods. One way to reduce computational costs is to approximate these functions using noisy estimates based on much smaller subsets of the data. The intuition here is that much of the data at hand is ``redundant'' for learning the posterior so it is more efficient from a computational and memory perspective to perform Bayesian inference on a subset of the data. For example, stochastic variational inference \citep{Hoffman:2013} uses minibatches of data to approximate gradients in a variational context. Stochastic gradient MCMC methods \citep{Ma:Chen:Fox:2015, WelTeh2011} perform a similar approximation in a gradient-based MCMC setting. In a Gaussian process context, as mentioned in Section~\ref{sec:sparse_local}, SVI has been used to speed up inference in sparse Gaussian processes from $O(NM^2)$ to $O\left(  M^2\min \left\{ M,B \right\} \right)$, where $B$ is the minibatch size.

An alternative is to use a minibatch to approximate the full posterior. Several embarrassingly parallel  MCMC methods combine noisy posterior estimates obtained using subsets of the data \citep{Minsker:2014}. \citet{SriCevDinDun2015} show that such stochastic approximation of sub-posteriors is strongly consistent. The Bayesian coresets approach aims to learn the posterior based on a reweighted posterior \citep{Huggins:2016}. While not directly equivalent (since it uses a single subset), \cite{Banerjee:2008} approximates a full Gaussian process model using a smaller subset of the data to form a prediction of the entire model. 

\section{Embarrassingly parallel inference with importance sampled mixture of experts}\label{sec:method}
We now introduce our novel method of fitting mixtures of Gaussian processes.
We assume our covariates $ X = \left\{ x_1, \ldots , x_N \right\} $ are distributed according to a Dirichlet mixture of $ K $ Gaussian components,
\begin{align}
\begin{split}
x_i \sim \mbox{Normal}(\mu_{z_i}, \Gamma_{z_i}), \hspace{1em} (\mu_k, \Gamma_k) \sim \mbox{Normal-Inv. Wishart}(\mu_0, \lambda, \Psi, \nu)\\
z_i \sim \mbox{Categorical}(\pi), \hspace{1em} \pi \sim \mbox{Dirichlet}(\alpha).
\end{split}\label{eqn:gmm}
\end{align}

The outputs are then assumed to be generated by $K$ independent Gaussian processes,
\begin{equation}
  \{y_i:z_i=k\}| \left( \{x_i:z_i=k\}, \Theta_k \right) \sim \mbox{GP}(0, \Sigma_{\Theta_k}),\label{eqn:ind_gps}
\end{equation}
where $\Sigma_{\Theta_k}$ is the covariance matrix between $\{x_i:z_i=k\}$ parametrized by $\Theta_k$. Conditioned on the $z_i$, we have a simple local GP with block diagonal structure, of the sort considered in Section~\ref{sec:sparse_local}. Rather than invert the $N\times N$ covariance matrix, we only need invert the $K$ blocks. These can be inverted in parallel using $K$ threads, each costing $O(N^3/K^2)$.

Marginalizing over the $z_i$ in Equation~\ref{eqn:gmm} to give a mixture-of-experts model  avoids the key limitation of the local GP methods: that they ignore correlation between the fixed blocks. In a setting like ours, where the inputs are clustered based on location, this means we ignore long-range correlation. Conversely, a mixture-of-experts approach allows long-range correlations and yields a dense expected covariance matrix.

The typical mixture of experts approach of marginalizing over partitions using MCMC is expensive (due to slow mixing) and difficult to parallelize. Instead, we use a trivially parallelizable importance sampling scheme. In short, we independently sample $J$ partitions of the input space, conditioned on the covariates $X$, by sampling from a Dirichlet mixture of $K$ Gaussians, conditioned on the inputs $X$ and ignoring the outputs $Y$, 
\begin{equation*}
 P(z_i =k | -) \propto \pi_{k}P(X_i |\mu_k, \Sigma_{k}),
\end{equation*}
where the mixture parameters are drawn from the prior distribution $ P(\mu,\Sigma) $, which we assume is Normal-Inverse Wishart. 

We then fit independent Gaussian processes to each of the $K$ partitions of each of the $J$ samples. We then (independently) calculate the appropriate importance weights, and use these weights to combine the $J$ samples. If our we assume the latent function we are modeling is generated from a mixture of GP functions, then our proposed method is an exact algorithm for fitting the model and calculating the marginal likelihood. However, if the latent function is a single GP then our algorithm is a fast approximation of the full GP and its marginal likelihood.

To further reduce memory and computational constraints, we can sample size $ B << N $ minibatches of the data without replacement and approximate the full likelihood by raising it to the $ N/B $ power. We detail these steps below in Sections~\ref{sec:is} and \ref{subsec:mb}, and provide a summary of the process in Algorithm~\ref{alg:ISMOE}.

\begin{algorithm}
	\caption{Importance Sampled Mixture of Experts (IS-MOE)}
	\label{alg:ISMOE}
	\For{$ j = 1 , \ldots , J $ in parallel}{
		Draw partition with $ K$ clusters  of data from $ P(Z|X)$\\
		Fit $ K $ independent GP models on the partitioned data.\\
		Predict new observations on each importance sample with \begin{equation*}P(f^{\ast}_{j} | Z_j, - ) = \sum_{k=1}^{K} P(f^{\ast}_{j} |  Z_j^{\ast}, - ) P(Z_j^{\ast}| -).\end{equation*}\\
		Obtain weights $ w_j = \prod_{k=1}^{K} P(Y_{k,j} | X_{k,j}, Z_j)$.
	} 
	Normalize weights, $ w_j := w_j/\sum_{j=1}^{J}w_j$.\\
	Average predictions using importance weights: $P(\bar{f}^{\ast} | - ) = \sum_{j=1}^{J} w_j	P(f^{\ast}_{j} | Z_j, - )$
\end{algorithm}


\subsection{Design choices}
Our proposed method makes a number of design choices, each of which carries important consequences for the performance of our algorithm which we will discuss in this section. In Equation~\ref{eqn:ind_gps}, we assume that each GP has its own set of hyperparameters, $ \Theta_k $. This allows us to capture a degree of non-stationarity and heteroscedasticity, for minimal additional cost. Alternatively, if we believe the model is stationary, we can share hyperparameters across partitions.

The Gaussian likelihood in Equation~\ref{eqn:gmm} is a design choice that is chosen to be appropriate in a wide range of settings. A mixture of Gaussians allows us to exploit correlations in the input location, and encourages preservation of short-range covariances though alternative likelihoods could also be used. In Section~\ref{sec:importance}, we will show that placing structure on the input space produces better results than simple uniform partitioning of the data or, at the very least, produces results that are not different than uniform partitioning when there is no structure in the input space.  We consider the general setting where each GP has its own set of hyperparameters, $ \Theta_k $. 

To avoid explicitly selecting the number of mixtures, $ K $, to use to model our input space, we may instead draw partitions from the Dirichlet process mixture model (DPMM) instead, as seen in \cite{Rasmussen:Ghahramani:2002,Meeds:Osindero:2006,Yuan:Neubauer:2009} and control the number of partitions via the concentration parameter, $ \alpha $. We choose a finite mixture model for two reasons. First, a Dirichlet distribution with $\alpha>1$ avoids the rich-get-richer behavior of the Dirichlet process, encouraging similarly sized clusters rather than one very large cluster. Second, a finite mixture model allows us to explicitly investigate the effect of increasing the number of clusters on the performance of our algorithm. However, the issue of selecting $ K $ is vital in practice. To address this problem, we could fit a mixture model on the data (or a subset, in ``big data'' cases) beforehand and empirically estimate $ K $ from this mixture model's posterior. Or, we could adopt a more systematic method of selecting $ K $ by using a Bayesian optimization method \citep[for example]{Snoek:Larochelle:Adams:2012} to explore the optimal number of partitions.

\subsection{Importance sampling}\label{sec:is}
We wish to capture posterior uncertainty about the partition and the associated covariance function $\boldsymbol{\Sigma}$, while ensuring our algorithm can be distributed. Importance sampling allows us to estimate the posterior expectations $E[g(f)]$ of some functional of $f$, such as the posterior predictive distribution, using an appropriately weighted collection of samples from some simpler distribution. Unlike MCMC, these samples can be collected independently, facilitating distributed computing.

We choose our proposal distribution over partitions to be the posterior distribution $P(Z|X)$ under the Gaussian mixture model given in Equation~\ref{eqn:gmm} conditioned on the input values $X$, ignoring the output values $Y$. We obtain approximate samples from this distribution by drawing mixture locations randomly from the prior, $ P(\theta_k, \Gamma_k) $ and assign data to clusters from $ P(z_{i} = k | -) $. After fitting the local GPs derived from this partition,  we then weight these particles using self-normalized importance sampled weights \footnote{Since we are working with self-normalized weights, the estimate has a bias of $O(1/J)$ \citep{Kong:1992}, but will often have a lower variance than the unbiased estimate obtained with $w_j=p(Z|X,Y)/p(Z|X)$, which involves calculating an intractable normalizing constant.}
$$w_j \propto \frac{p(Z|X,Y)}{p(Z|X)},$$
where $\sum_j w_j = 1$ and $p(Z|X,Y) \propto P(Y,X|Z)P(Z)$ is the cluster assignments using output and input data (as opposed to only input data in $ P(Z|X)$). We can then obtain an asymptotically unbiased estimate $\hat{\mu}$ to $E[g(f)]$ as
\begin{equation*}
\hat{\mu} = \sum_{j=1}^J w_j g(f_j).
\end{equation*}
As a concrete example, the posterior predictive distribution is approximated as
\begin{equation*}
\hat{p}(f^*|x^*, X,Y) = \sum_{j=1}^{J} w_j p(f^*_j|Z_j,x^*,X,Y),
\end{equation*}
where $Z_j$ is the partition associated with the $j$th sample.

Calculating the $w_j$ involves integrating over the covariance parameters $\Theta$,
\begin{align}
  w_j\propto \frac{p(Z|X,Y)}{p(Z|X)} \propto \frac{p(X,Y|Z)p(Z)}{p(X|Z)p(Z)}= p(Y|X,Z) = \int p(Y|X,Z,\Theta)p(\Theta)d\Theta.
\end{align}
where $p(\Theta)$ is the prior over the covariance parameters. If we are allowing separate hyperparameters, $ \Theta_{j,k} $, for each partition, we assume that $p(\Theta) = \prod_kp(\Theta_k)$, so
\begin{align}
  w_j \propto& \prod_{k=1}^K p(\{y_i: z_{j,i} = k\}|\{x_i: z_{j,i} = k\})\\ =& \prod_{k=1}^K \int p(\{y_i: z_{j,i} = k\}|\{x_i: z_{j,i} = k\}, \Theta_{k})p(\Theta_k)d\Theta_k\end{align}

We must approximate the intractable integral. Depending on our accuracy/speed trade-off, we can obtain an unbiased estimate of the $w_j$ using a sample-based approximation; we can perform a Laplace approximation about the MAP solution $\hat{\Theta}$; or we can directly use the MAP approximation $p(Y|X,Z) \approx p(Y|X,Z,\hat{\Theta})$. In our experiments, we choose to directly use the MAP solution, obtained using gradient descent; while this is not as accurate as sampling hyperparameters it is significantly faster, and mirrors the choices made by our comparison methods.


Calculating the MAP approximation of $\Theta$ (or indeed, inferring the hyperparameters using MCMC or another method) requires calculating the marginal likelihood $p(Y|X,Z,\hat{\Theta})$. This means there is no additional cost involved in calculating the importance weights, up to a normalizing constant. Independence between each importance sample means that the samples and their normalizing constants can be obtained in parallel. The only global communication required is at the end of the procedure,  when the importance weights are normalized and the samples are combined to give our predictive distribution (or other desired expectation). In the regression scenario, we can obtain the exact marginal likelihood to fit mixtures of GPs due to the tractability of the classic regression model. However, we can show that in Section~\ref{sec:classification} it is possible to use an approximation to the marginal likelihood as an importance weight to fit classification IS-MOE models that obtain good performance.

 
The overall computational cost of the IS-MOE, using $J$ importance-weighted samples and $K$ blocks, is therefore $O(JN^3/K^2)$. In Table~\ref{table:complexity}, we compare this with the overall computational cost of the full GP, sparse approximations (FITC, DTC and SVI), the Bayesian treed GP (BTGP), and the robust Bayesian committee machine (RBCM). While the $O(JN^3/K^2)$ cost is $O(J)$ higher than sparse methods and local methods based on a fixed partition such as RBCM, we note that the $J$ samples can be performed and weighted in parallel---meaning the time taken is comparable if we are willing to sacrifice computational resources.\footnote{In general, a sparse model with $M$ inducing points obtains comparable accuracy to a local method with $N/K$ local GPs, and has equivalent computational complexity.} In this procedure, the only communication between processors occurs at the end of the prediction step when we normalize the weights, $ w_j $, and obtain the importance averaged predictions, $ \bar{f}^{\ast} $. This is vital in any distributed computation algorithm due to the high overhead cost of inter-processor communication. We can also make use of the independence of the $K$ partitions to parallelize further, using $JK$ threads each taking $O(N^3/K^3)$. As shown in Table~\ref{table:complexity}, this leads to an equivalent wall-time cost comparable with the distributed RBCM. As we will see in Section~\ref{sec:results}, the extra computational cost required to ensure a full posterior predictive distribution yields improved performance over methods that are based on a fixed partition.

However, importance sampling has inherent issues that can hinder practical performance for inference and prediction. First, importance samplers have a tendency to produce weights where one proposal completely dominates the rest of the proposals and obtains an importance probability of nearly one. To this end, we could smooth out the importance weights with a Pareto distributed smoother in order to obtain more stable estimates from our importance sampler \citep{Vehtari:Gelman:Gabry:2015}. Additionally, choosing a good proposal distribution is critical to the performance of the importance sampler but it is not obvious how to select the best distribution. \cite{Kahn:Marshall:1953} show that the optimal distribution which minimizes the estimator variance is $ |g(f)|p(f)$ for the expectation $ \mu = \int g(f)p(f)/q(f) \; df $ though practically speaking we may not be able to easily sample from this distribution or calculate importance weights.

\subsection{Minibatched importance samples}\label{subsec:mb}
Although we can obtain significant computational and memory saving advantages using our low-rank approximation, we still may encounter major bottlenecks from attempting to approximate the covariance matrix of the full training set. To overcome this issue, we propose a ``minibatching'' solution, where each importance sample is obtained and weighted based only on a subset of size $B<<N$ sampled uniformly without replacement. Given a random subset $B$ of observations, we can approximate $p(\Theta|X,Y,Z)$ with the subset posterior $p(\Theta|X^{\small{mb}},Y^{\small{mb}}, Z)$ evaluated on a size-$B$ minibatch $(X^{\small{mb}},Y^{\small{mb}})$. Such a posterior estimate is strongly consistent, but will tend to underestimate the posterior variance \citep{SriCevDinDun2015}. To achieve realistic credible intervals, we can assume we have seen each pair $(x_i, y_i)$ in our minibatch $N/B$ times; mathematically, this corresponds to raising the contribution of the likelihood to the subset posterior to the $(N/B)$-th power. 

We use this stochastic approximation trick to estimate the posterior distribution over parameters for each importance sample, allowing us to reduce our overall complexity from $O(JN^3/K^2)$ to $O(JB^3/K^2)$. Empirical results in Section~\ref{sec:results} will show that this stochastic approximation performs favorably on large datasets in comparison with both the non-SA IS-MOE method and other scalable GP inference methods.

\begin{table}[t]
  \caption{Comparison of inference complexity. $N$ is the number of data points, $K$ is the number of experts or local GPs, and $M = N/K$ is the number of inducing points. For the Monte Carlo based methods, $J$ is the number of MCMC iterations or importance samples.}
  \label{table:complexity}
  \centering
      \begin{tabular}{lcccccccccccc}
    
    & \multicolumn{4}{c}{Full GP} & \multicolumn{4}{c}{Sparse} & \multicolumn{4}{c}{SVI}  \\
    \midrule
    Complexity & \multicolumn{4}{c}{$N^3$} & \multicolumn{4}{c}{$NM^2$} & \multicolumn{4}{c}{$M^2\min(M,B)$}  \\ 
    & & & & &&&&&&&&\\
    & \multicolumn{3}{c}{RBCM} & \multicolumn{3}{c}{BTGP} &  \multicolumn{3}{c}{\textbf{IS-MOE}} & \multicolumn{3}{c}{\textbf{SA IS-MOE}}  \\
    \midrule
    Complexity  &\multicolumn{3}{c}{$N^3/K^2$} & \multicolumn{3}{c}{$JN^3/K^2$}  & \multicolumn{3}{c}{$JN^3/K^2$} & \multicolumn{3}{c}{$JB^3/K^2$} \\
    Comp/thread &\multicolumn{3}{c}{$N^{3}/K^{3}$} & \multicolumn{3}{c}{$\times$} & \multicolumn{3}{c}{$N^3/K^3$} & \multicolumn{3}{c}{$B^3/K^3$} \\
  \end{tabular}
\end{table}\

\section{Experimental evaluation}\label{sec:results}
To showcase the performance of our method, we compare it with a number of competing methods on both synthetic and real data sets.

\subsection{Evaluation on synthetic data}\label{sec:synth}

\subsubsection{Comparison with competing methods}
We begin by evaluating our method on synthetically generated data, in order to allow us to explore and visualize a range of regimes, and to allow comparison with methods that do not scale to our real-world dataset. In our studies, we will compare our Importance Sampled Mixture of Experts approach (IS-MOE) against a full Gaussian process (GP); three sparse approximations to this model: FITC~\citep{Snelson:Ghahramani:2005}, DTC~\citep{Seeger:Williams:Lawrence:2003}, and SVI~\citep{Hensman:2013}; the Bayesian treed GP \citep[BTGP]{Gramacy:Lee:2005}; and the robust Bayesian committee machine \citep[RBCM]{Deisenroth:Ng:2015}. All models use a squared exponential covariance matrix. Our IS-MOE code uses the Gaussian process modules in \texttt{GPy} \nocite{gpy:2014} in Python with parallelization executed through \texttt{mpi4py} \citep{Dalcin:2005}.\footnote{The code is available at \url{https://github.com/michaelzhang01/ISMOE}.} We ran the full GP, FITC, DTC and SVI implementations also through \texttt{GPy}, BTGP in \texttt{tgp}, and RBCM in \texttt{gptf}. 

We first consider three data settings, the first two were generated on a linearly spaced grid of values on $[-1,1]$ and the last one has more data generated in the center of the range $[-1,1]$ in order to test how our method approach works when data are not uniformly generated. For these experiments we did not use minibatching in IS-MOE.
\begin{enumerate}
\item \textbf{Stationary, long-range correlations} generated with inverse length scale $ \gamma=15 $.\label{data1}
\item \textbf{Stationary, short-range correlations} generated with inverse length scale $ \gamma=5000 $.\label{data2}
\item \textbf{Non-stationary} generated piecewise with fast and slow moving periodic functions.\label{data3}
\end{enumerate}

In examples \ref{data1} and \ref{data2}, we generated data from a GP with zero mean squared exponential covariance kernel with amplitude $ \nu = 1 $. For all examples we added Gaussian noise $ \sigma^{2}=1 $ to the observed outputs. We generated a training data set with 1,000 observations and a test set with 100 observations.

\begin{figure}
	\centering
	\includegraphics[width=.9\textwidth]{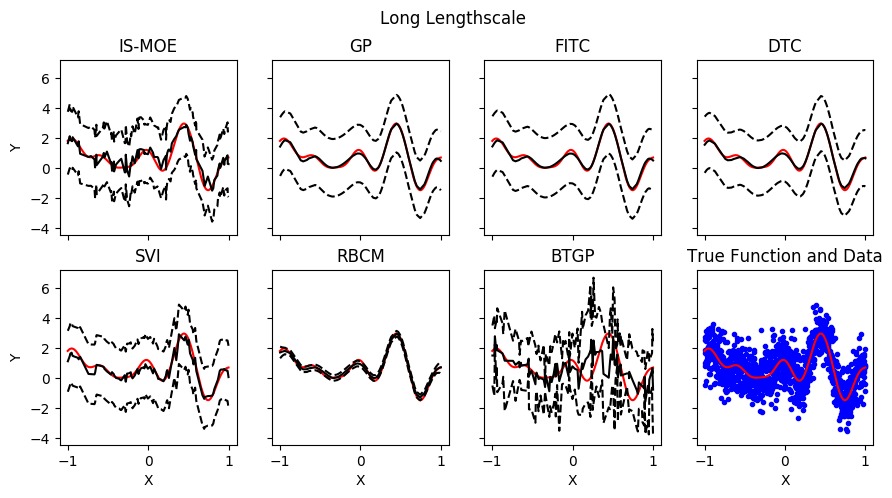}
	\caption{Posterior mean and $95\%$ predictive intervals on Synthetic 1 (stationary, long length scale).}
	\label{fig:long_ls}
\end{figure}

\begin{figure}
	\centering
	\includegraphics[width=.9\textwidth]{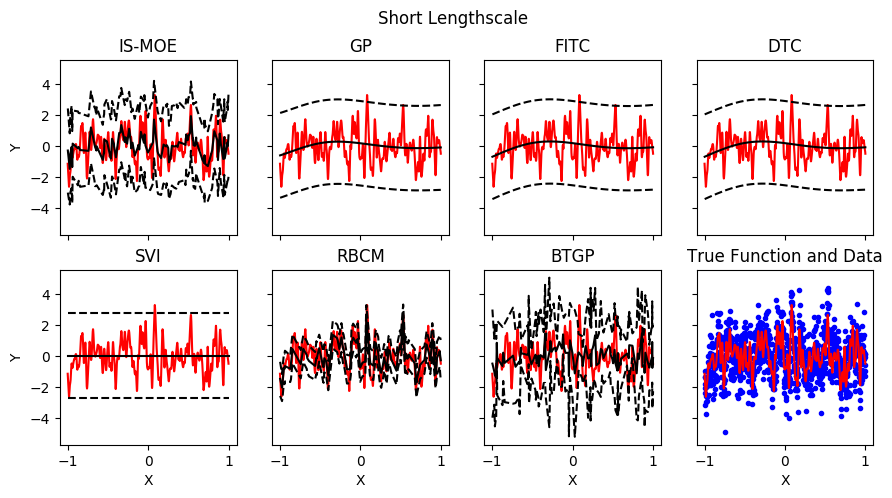}
	\caption{Posterior mean and $95\%$ predictive intervals on Synthetic 2 (stationary, short length scale).}
	\label{fig:short_ls}
\end{figure}

\begin{figure}
	\centering
	\includegraphics[width=.9\textwidth]{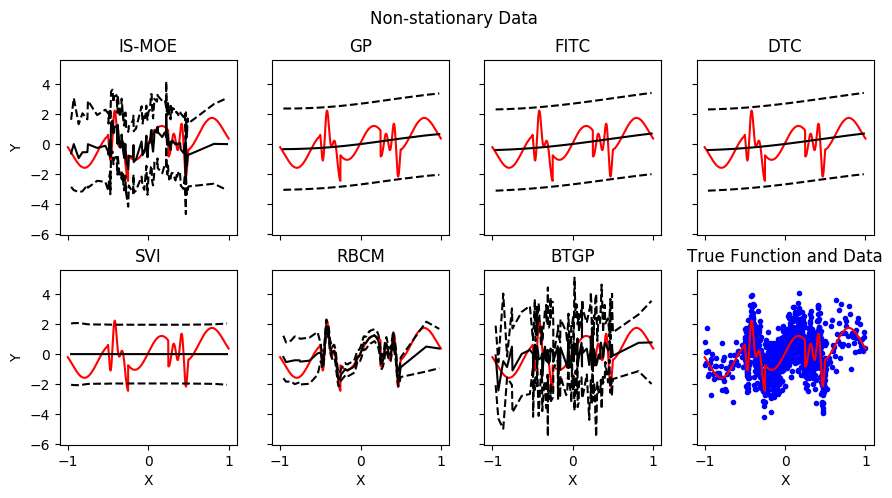}
	\caption{Posterior mean and $95\%$ predictive intervals on Synthetic 3 (non-stationary).}
	\label{fig:nonstationary}
\end{figure}

For fitting the stationary data, we restrict the hyperparameters on our IS-MOE method to be the same on all $ K $ blocks. For (\ref{data3}), we allowed each mixture to have its own hyperparameters in order to model the non-stationarity of the data. In all methods except BTGP we infer hyperparameters through the MAP estimate via gradient descent optimization, and for BTGP we infer the hyperparameters through MCMC sampling. For the sparse methods, we used $M=100$ inducing points, and for the local methods (including the IS-MOE) we used $K=10$ partitions to have a comparable level of computational complexity. For the BTGP we ran the MCMC sampler for $ 10 $ iterations; for the IS-MOE we used $J= 10 $ independent importance-weighted samples. Figures~\ref{fig:long_ls}, \ref{fig:short_ls}, and \ref{fig:nonstationary} shows the posterior predictive results and predictive intervals obtained using the five methods, and Tables~\ref{table:regression_ll} and \ref{table:regression_mse} show the corresponding test set log likelihoods and mean squared errors.

We first consider the one-dimensional stationary examples. Recall that, in general, sparse methods perform well when the covariance structure is dominated by longer-range correlations, and local methods perform well when we have significant local variation in our function. For these results, we deliberately set $ J $ to a small number to see how IS-MOE performs when there is ``not enough'' importance samples as a difficult scenario in comparison to the other methods. Looking at the results on the dataset with long-range correlations (Figure~\ref{fig:long_ls}), we see that the IS-MOE can capture the predictive variance unlike RBCM which is over-confident in its predictions, and only performs slightly worse than the full GP and the sparse approximations, again due to a small number of importance samples. 

If we look at the dataset with short-range correlation (Figure~\ref{fig:short_ls}), we see the sparse methods struggle to learn the function--with a small number of inducing points, it is impossible to capture the high-frequency variation. Looking at the quantitative results in Tables~\ref{table:regression_ll} and \ref{table:regression_mse}, we see that the IS-MOE outperforms the RBCM because our method is capable of learning the proper predictive variance whereas the RBCM is over-confident in its results. The BTGP likely produces poor predictive performance is because of lack of convergence of the MCMC chain: the underlying model is fairly complex and will tend to mix slowly and at a comparable level of computational complexity in this evaluation, does not have enough MCMC iterations to converge. 

\begin{figure}
	\centering
	\includegraphics[width=1.0\linewidth]{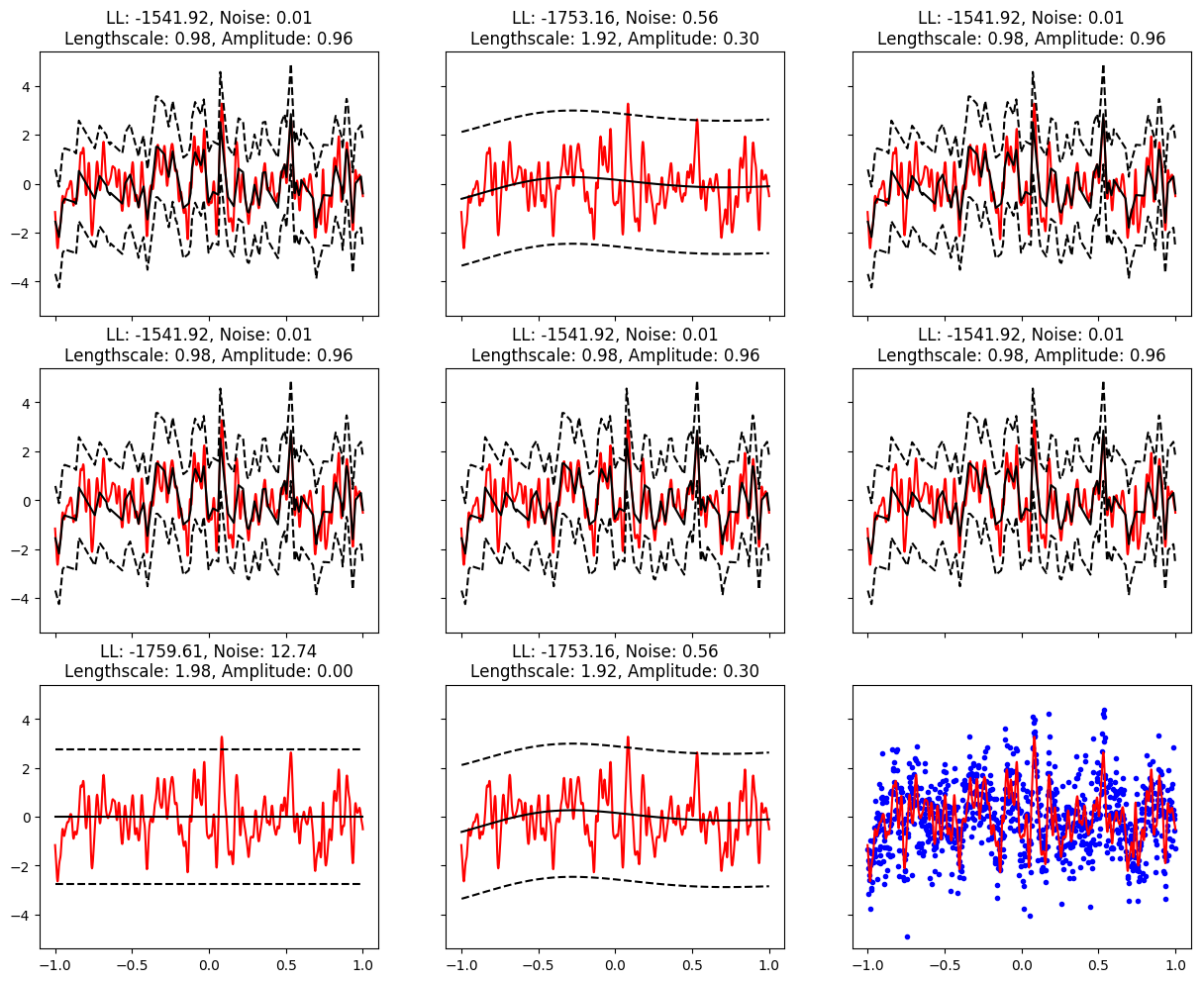}
	\caption{Multiple instantiations of full Gaussian process fits to short lengthscale data. The true latent function is plotted in red with the predictive mean in black with $ 95\%$ credible intervals in black dashed lines. The observed data is plotted in blue dots.}\label{eqn:multiple_gp_short_ls}
\end{figure}

We can also see that the full GP struggles in this scenario between assuming the function is one that exhibits high noise in the data or that the function exhibits short-range correlation. Figure~\ref{eqn:multiple_gp_short_ls} shows an example of this multimodal structure in the marginal GP likelihood. Here, if we randomly initialize the hyper-parameter values from the distribution $ \mbox{Gamma}(1,4) $ for the full GP we can see that the optimizer can get stuck in suboptimal modes. Monte Carlo methods like importance sampling generally do not suffer from converging to suboptimal local optima as much as gradient descent methods or MCMC would because importance samplers will independently explore different parts of the marginal likelihood surface and place more weight on particles in more optimal modes than less optimal ones.

Finally, consider the non-stationary example, which combines known failure modes of local and sparse GPs. We have a combination of slowly varying behavior (which is poorly captured by local methods) and fast-varying behavior (which is poorly captured by sparse methods). The full GP, RBCM and sparse methods, fitted with stationary kernels, obviously cannot account for the non-stationary components in the data, and by assuming a stationary covariance they give poorer test-set performance. The BTGP does a reasonable job at capturing the function; again its performance is likely to be hampered by slow mixing and lack of convergence. Figure~\ref{fig:nonstationary} shows that the IS-MOE is able to capture the function, and Tables~\ref{table:regression_ll} and \ref{table:regression_mse} show that it can provide confident predictions at all regions of the function.  

\begin{table*}
	\caption{Test set performance on synthetic datasets}
	\begin{subtable}[b]{\linewidth}
		\centering
		\caption{Log likelihood}
		\begin{tabular}{lrrrrrrr}
			\toprule
			Data &	\textbf{IS-MOE} &       GP &     FITC &      DTC &      SVI &      RBCM &     BTGP \\
			\midrule
			Long Lengthscale  & -152.41 & -143.52 & -143.39 & -143.81 & -156.51 & -5207.89  & -231.84  \\	
			Short Lengthscale & -157.16 & -172.32  & -172.26 & -172.26 & -173.38 &  -251.50 & -212.61 \\
			Non-stationary    & -158.21 & -181.40 & -181.30   & -181.30   & -198.00 &  -910.54 & -256.73 \\
			\bottomrule
		\end{tabular}
		\label{table:regression_ll}
	\end{subtable}
	\vspace{0.05in}
	
	\begin{subtable}[b]{\linewidth}
		
		\caption{MSE}
		\centering
		
		\begin{tabular}{lrrrrrrr}
			\toprule
			Data              &    \textbf{IS-MOE} &      GP &    FITC &     DTC &     SVI &    RBCM &    BTGP \\
			\midrule
			Long Lengthscale  & 1.20 & 1.03  & 1.02 & 1.02 & 1.30 & 1.03 & 2.07 \\
			Short Lengthscale & 1.35 & 1.83 & 1.83 & 1.83 & 1.87 & 1.06 & 2.46 \\
			Non-stationary    & 1.39 & 2.18 & 2.17 & 2.17 & 2.12 & 1.00 & 2.64  \\
			\bottomrule
		\end{tabular}
		\label{table:regression_mse}
	\end{subtable}
\end{table*}
\begin{figure}
	\centering
	\includegraphics[width=.90\linewidth]{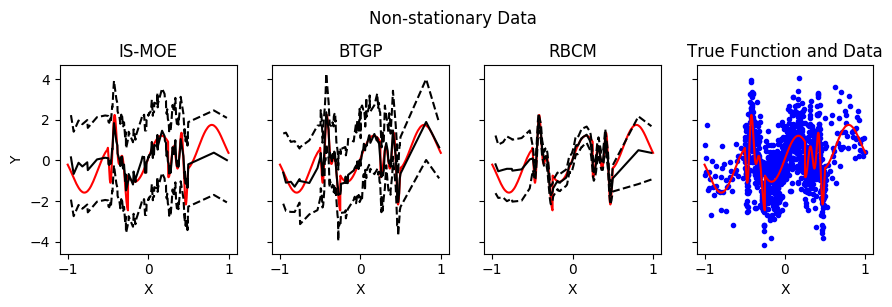}
	\caption{Posterior mean and $95\%$ predictive intervals on Synthetic 3 (non-stationary), $ J=6000 $.}	\label{fig:nonstationary5000}
\end{figure}
\begin{table}
	\centering
	\caption{Predictive log likelihood, MSE and total run time for the non-stationary example, $ J=6000 $}
	\begin{tabular}{lrrr}
	\hline 
	& Log Likelihood & MSE & Wall time (s.)\\
	\hline
	\textbf{IS-MOE} & -154.79 & 1.28 & 61.16\\
	BTGP & -145.71 & 1.04 & 1116.75 \\ 
	RBCM & -910.55 & 1.01 & 13.08\\
	\hline
	\end{tabular}
	\label{table:ns}
\end{table}
To provide a deeper comparison with the treed Gaussian process, we then run the non-stationary example with the local methods (IS-MOE, BTGP and RBCM, as these methods are the only ones considered in the experiments that can model non-stationary data) when there is sufficient computational resources for the MCMC chain in the BTGP to converge (or in other words, when we set $ J = 6000 $). In Figure~\ref{fig:nonstationary5000}, we see that BTGP learns a much smoother latent function that the one in Figure~\ref{fig:nonstationary} while the results for IS-MOE and RBCM still largely remain the same. Table~\ref{table:ns} shows the predictive and timing comparisons between the three methods. We see that BTGP obtains the best predictive log likelihood and RBCM obtains the best MSE and fastest run time. However, our method, IS-MOE, produces the best compromise between BTGP and RBCM as we obtain only slightly worse predictive log likelihood results than BTGP while being more than eighteen times as fast. Thus, we can understand our method as being a fast, parallelizable, approximate variant of the BTGP. Though IS-MOE is slower and less accurate (with respect to MSE) than RBCM in this experiment, we obtain far better uncertainty quantification of our predictions as reflected in the poor log likelihood results of the RBCM.

\subsubsection{The importance of importance sampling}\label{sec:importance}
\begin{figure}[t]
	\centering
	\includegraphics[width=.8\linewidth]{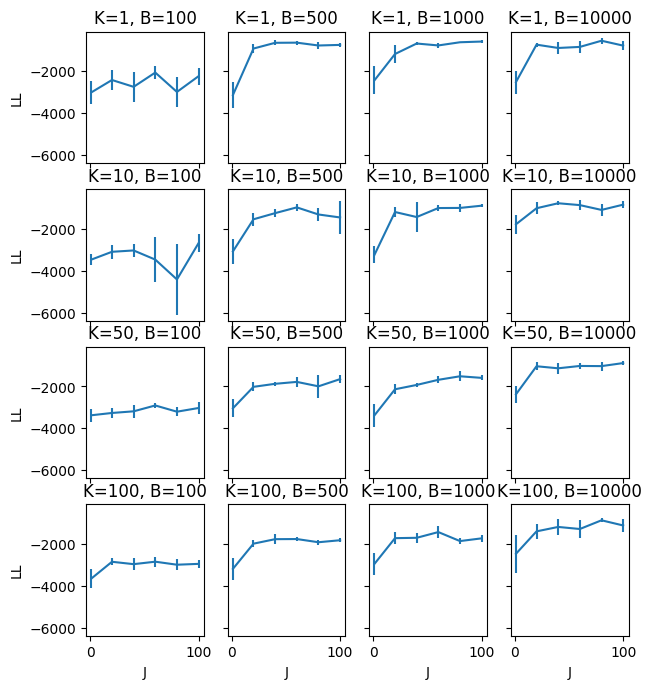}\\
	\caption{Evaluation on synthetic data of the effect of the number of samples $J$ on the test set log likelihood of IS-MOE (with 95\% confidence intervals), for various values of $B$ and $K$.}
	\label{fig:njk_synthetic}
\end{figure}
\begin{figure}[ht]
	\centering
	\includegraphics[width=.8\linewidth]{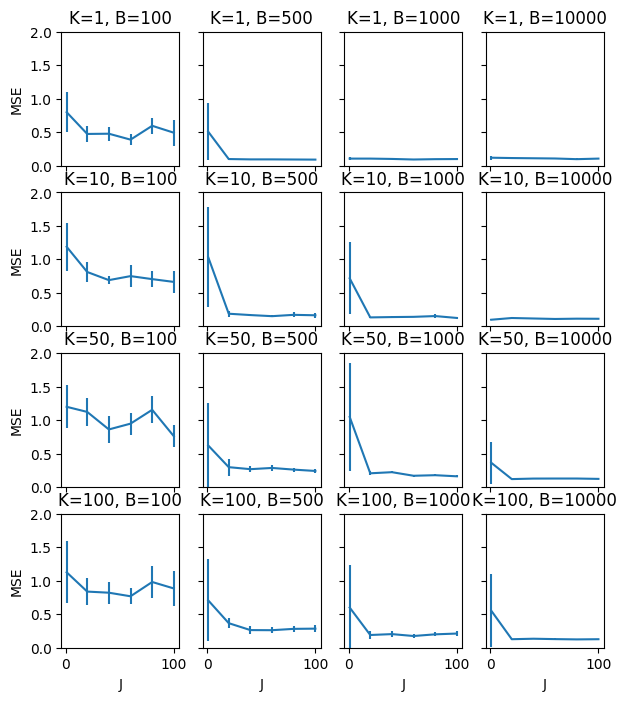}
	\caption{Evaluation on synthetic data of the effect of the number of samples $J$ on the test set MSE of IS-MOE (with 95\% confidence intervals), for various values of $B$ and $K$.}
	\label{fig:njk_synthetic_mse}
\end{figure}
The IS-MOE falls under the ``local'' framework, much like the RBCM and the BTGP; however it out-performs both methods. This can be attributed to importance sampling a distribution over partitions. To demonstrate this, we consider our performance on a synthetic dataset of 10,000 training observations, 2,000 test observations and 100 covariates. The input data are sampled from a GMM on $\mathbb{R}^{100} $  with $K= 50$ mixture components. The  output, $ Y $, is drawn from a GP model with zero mean and an RBF kernel with inverse length scale $.001$, observation noise variance of $ .25 $ and amplitude of $2$.

\begin{figure}[ht]
	\centering
		\includegraphics[width=.9\linewidth]{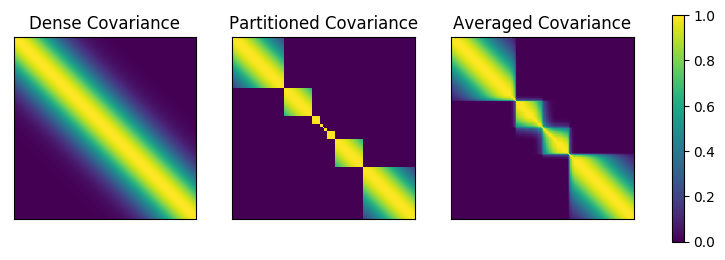}
	\caption{Dense, partitioned and averaged covariance matrices for a latent function with a short lengthscale. ``Dense Covariance'' is the dense covariance matrix $ \Sigma(X,X^{\prime}) $.  ``Partitioned Covariance'' is one instance of a block diagonal covariance matrix. ``Averaged Covariance'' is an averaging of several block diagonal partitioned covariance matrices.}\label{fig:cov}
\end{figure}

The RBCM uses a single, fixed partition. Conversely, the IS-MOE uses a distribution over partitions, combined using importance sampling weights. Figures~\ref{fig:njk_synthetic} and~\ref{fig:njk_synthetic_mse} show how varying the number of importance samples, for a range of values of $K$ and $B$ (remember, $K=1, B=N=10,000$ corresponds to the full Gaussian process, and as $K$ increases or $B$ decreases, we expect a drop in quality). In most cases, we see a similar pattern: there is a clear improvement in performance between $J=1$ to around $J=50$, but beyond that the improvements level off. This confirms that averaging over partitions improves performance, but suggests that in this setting, we need relatively few samples to approximate the posterior. In Figure~\ref{fig:cov}, we can visualize why this is the case if we compare the resulting expected covariance matrices in a product of expert type approach like the RBCM with a mixture of expert approach like ours. We note that BTGP also averages over partitions and can achieve high quality predictions as a result; however the slow mixing of the MCMC algorithm and the inability to distribute inference means we get worse performance for the same computational effort, and precludes the use of BTGP on large datasets.

The IS-MOE uses importance sampled weights to average over partitions. In the minibatch setting, these weights and the samples themselves are obtained using a stochastic approximation. It is reasonable to question whether either the calculation of importance weights, or the up-weighting of the likelihood to obtain a stochastic approximation, affect the performance. In other words -- would we do as well using uniform weights or avoiding the stochastic approximation? As we see in Table~\ref{table:weight_compare} (which uses the same synthetic dataset as above, with $J=10$, $K=10$ and $B=1000$), using importance samples with reweighted likelihood minibatches results in better predictive performance than either not upweighting the minibatched likelihood or using uniform weights to combine predictions.

\begin{table}
		\caption{Test set log likelihood and MSE for various weighting schemes. Standard errors are in parentheses}
	\centering
	\begin{tabular}{lrr}
		\toprule
		Setting           &       LL &    MSE \\
		\midrule
		IS with SA      & -354.96 (28.59) & 0.096 (0.003) \\
		IS without SA   & -423.93 (41.24) & 0.097 (0.002) \\
		Unif. with SA    & -726.67 (14.19) & 0.19 (0.019)  \\
		Unif. without SA & -842.88 (9.82) & 0.25 (0.337)  \\
		\bottomrule
	\end{tabular}
	\label{table:weight_compare}
\end{table}

\begin{table}
  \caption{Test set log likelihood and MSE for two different covariate partitioning schemes when the input space is generated according to a GMM, and where the inputs are i.i.d Uniform.  Standard errors are in parentheses.}
  \begin{tabular}{llrr}
    \toprule
    Generating mechanism     & IS-MOE partitioning scheme & LL              & MSE \\
    \midrule
    \multirow{2}{*}{GMM}     & GMM                        & -429.19 (41.57) & 0.14 (0.01) \\
    & Random Clusters            & -789.37 (37.43) & 0.53 (0.02) \\
    \midrule
    \multirow{2}{*}{Uniform} & GMM                        & -2962.01 (6.45) & 0.62 (0.01) \\
    & Random Clusters            & -2964.29 (7.00) & 0.62 (0.01) \\
    \bottomrule
  \end{tabular}
  \label{table:gmm_unif_compare}
\end{table}
A final difference from the RBCM is the choice of the distribution over partitions that the IS-MOE is able to explore. The IS-MOE uses a distribution based on covariate location, while the RBCM generates its single partition uniformly. To evaluate the impact of using covariate location to guide the partitioning, we explore two variants of the IS-MOE model: one that uses a Gaussian mixture model to partition data, and one which uses random, uniform partitions. We tested these two methods on two synthetic datasets. The first was the synthetic dataset described above, with 100-dimensional inputs sampled from a mixture of 50 Gaussians. The second used the same kernel, but used 100-dimensional inputs with each dimension sampled uniformly from a $ \mbox{Uniform}(-1,1) $ distribution. For both cases, our IS-MOE model used  $J=10$, $K=10$ and $B=1000$; note that the number of clusters used does not match the number of generating clusters.  As we can see in Table~\ref{table:gmm_unif_compare}, when the input data exhibits clustering behavior, we perform better when we place structure in the input clustering as opposed to purely random partitioning. When there is no structure (i.e.\ the inputs are sampled from a uniform distribution) there is no significant difference between the two models. This suggests that using a GMM is a reasonable default choice, since it can take advantage of any structure in the input space but does not degrade performance in the absense of structure.

\begin{figure}
	\centering
	\includegraphics[width=1.0\linewidth]{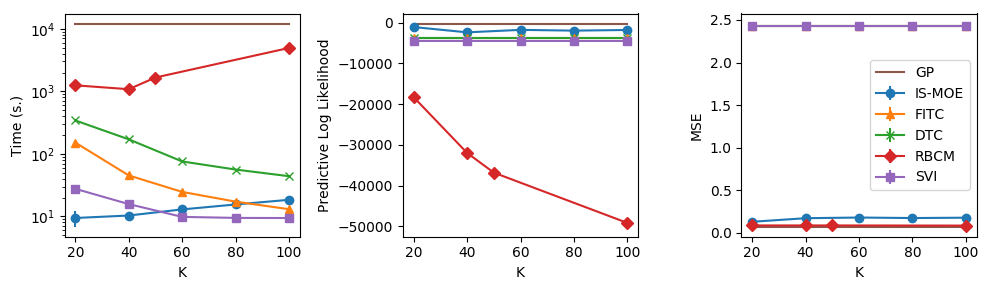}
	\caption{Wall clock time, predictive log likelihood and MSE results for large synthetic dataset with increasing $ K $. SVI, FITC and DTC obtain very similar predictive performances.}\label{fig:timing}.
\end{figure}

Lastly, we are interested to see how different GP methods compare with regards to the total wall clock time for fitting the model. Using the same data set, with $ J = 128$ and $ B=1000 $ for SVI and IS-MOE, we evaluate the speed and predictive performance of IS-MOE against the other methods compared in this paper. As seen in Figure~\ref{fig:timing}, we can see that IS-MOE is faster than all other methods (most notably, SVI) at low values of $ K $ while maintaining good predictive log likelihood and MSE performance. Only RBCM can outperform IS-MOE compared to the full GP in terms of MSE, but, as cited earlier, RBCM will systematically produce overconfident predictions. Counter-intuitively, we would expect local methods to perform faster with respect to wall time as $ K $ increases, but the results of IS-MOE and RBCM show that increasing $ K $ actually can increase wall time. We claim that the reason for this is because there is additional computational overhead involved with increasing the number of experts, as we must increase the number of GP models instantiated, whereas increasing $ K $ for sparse methods corresponds to reducing $ M $, meaning there are less parameters to fit in the sparse model. 

\subsection{Evaluation on real data}
As seen in our experiments on synthetic data, the IS-MOE is applicable to many different data regimes where other approximations may fail. Its inherently parallelizable nature also makes it an appealing choice for larger, real-world datasets where use of a full GP is computationally infeasible. To evaluate performance in this ``big data'' regime, we used an empirical dataset consisting of 209,631 mid-tropospheric CO2 measurements over space and time from the Atmospheric Infrared Sounder (AIRS)\footnote{Available in the R package \texttt{FRK} as \texttt{AIRS\_05\_2003}}. First, in Section~\ref{sec:sensitivity}, we use this dataset to explore the sensitivity of our model to different parameter values, to show how we can trade off between predictive accuracy and computational cost. Then, in Section~\ref{sec:comparison_real} we compare its performance against competing approaches.

\subsubsection{Sensitivity to model settings}\label{sec:sensitivity}
\begin{figure}[ht]
	\centering
	\includegraphics[width=.8\linewidth]{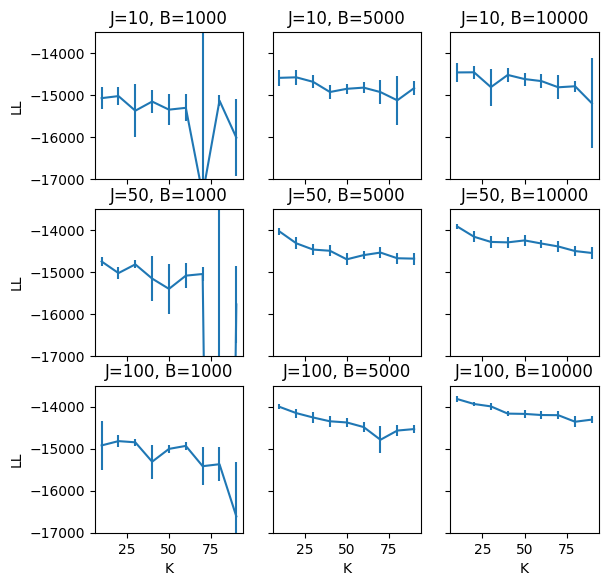}\\
    \caption{Evaluation on the AIRS dataset of the effect of $B$, $K$, and $J$ on the test set log likelihood of IS-MOE (with 95\% confidence intervals).}\label{fig:njk_test}
\end{figure}
\begin{figure}[ht]
	\centering
	\includegraphics[width=.8\linewidth]{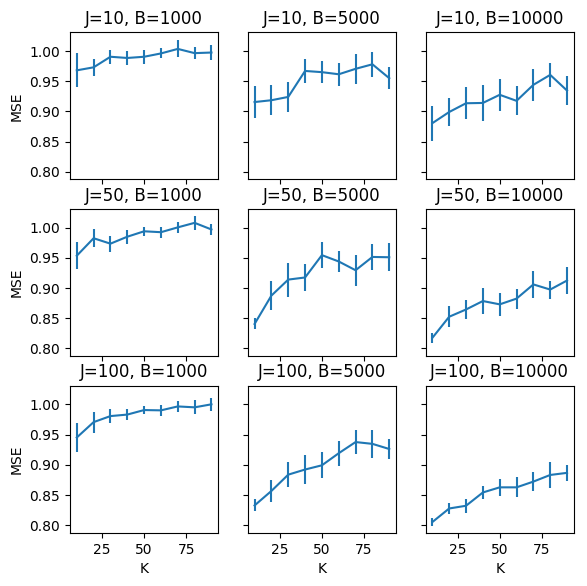}
    \caption{Evaluation on the AIRS dataset of the effect of $B$, $K$, and $J$ on the test set MSE of IS-MOE (with 95\% confidence intervals).}\label{fig:njk_test_mse}
\end{figure}
Clearly, both the performance and the cost of inference of our model will depend on the number of blocks $K$ in our approximation, the number of importance samples $J$, and the minibatch size $B$. On the one hand, inference scales as $O(JB^3/K^2)$, so we can speed up inference by decreasing $J$ or $B$ or by increasing $K$. On the other hand, a smaller number of blocks will allow us to better approximate a dense covariance matrix; a larger number of importance samples helps us explore the full posterior; larger minibatches reduce the noise in our estimators.

In order to pick values for $K$, $J$ and $B$, we must understand how they affect our overall estimates. We trained the IS-MOE using  a range of values for $ B $, $ K $ and $ J $, over $ 20 $ cross-validation splits. 
As expected, we find that as we increase $K$ or decrease $J$ and $B$ our performance deteriorates. Figure~\ref{fig:njk_test} shows that as the $ B $ and $ J $ increases, the average predictive log likelihood increases and the variance of the log likelihood decreases, and that as $ K $ increases the quality of our inference method degrades. However, looking at Figure~\ref{fig:njk_test}, we see the deterioration in predictive likelihood is fairly gradual for most values: we only see a dramatic degradation when we have both a small minibatch size and a large number of partitions. This suggests that the practitioner can modify $B$, $J$ and $K$ within a wide range to achieve acceptable computational costs without a dramatic drop in quality.

\subsubsection{Comparison with competing methods}\label{sec:comparison_real}
\begin{figure}
	\includegraphics[width=.5\linewidth]{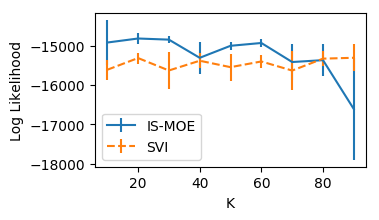}\includegraphics[width=.5\linewidth]{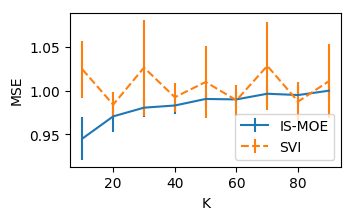}\\
	\caption{Comparison of IS-MOE and SVI on AIRS dataset for various $K$, with $ J=100 $ and $ B=1000 $. SVI parameters chosen to have equivalent computational cost.}\label{fig:big_data_compare}
\end{figure}

Using the same CO2 dataset and squared exponential kernel as before, we compare the IS-MOE with SVI -- the only other method that would scale to this dataset.\footnote{While RBCM is designed to scale to large data, we were unable to run the available Python package \texttt{gptf} due to memory issues.} For IS-MOE, we set $J=100$ and $B=1000$ and explored a range of values of $K$; for SVI we chose values for inducing points that gave a comparable level of computational complexity. We evaluated performance over 20 cross-validation splits. Our importance sampling method provides for a richer predictive model due to the averaging over importance proposals, and we see the benefit of this in our results. As Figure~\ref{fig:big_data_compare} shows, the IS-MOE typically performs comparably to SVI at equal levels of computational complexity in predictive performances using both metrics until approximately 90 clusters, in which our method performs notably worse due to the long range correlation present in this dataset. 

\subsubsection{Applications beyond regression}\label{sec:classification}
Finally, to highlight that our method is not limited to a specific GP model, we apply our method on a binary classification task, using a Laplace approximation with a squared exponential kernel. We compare with the full GP \citep{Williams:Barber:1998} and the sparse GP \citep{Hernandez-Lobato:2016} on three classification datasets from the UCI repository: the Pima Indians diabetes dataset; the Parkinsons dataset; and the Wisconsin diagnostic breast cancer (WDBC) dataset.\footnote{All empirical classification datasets are available in the UCI repository at \texttt{http://archive.ics.uci.edu/ml/}.} As Table~\ref{table:classificationLL} shows, the IS-MOE can approximate the full GP results very well, with comparable area under the curve (AUC) scores and log likelihood to a full GP and a sparse approximation.

\begin{figure}
	\centering
	\includegraphics[width=1.0\linewidth]{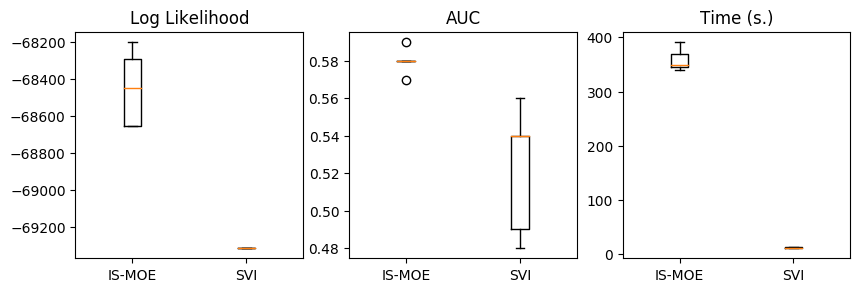}
	\caption{Results for Higgs-Boson classification experiment.}
	\label{fig:higgs}
\end{figure}

In addition, we also ran IS-MOE in comparison with the sparse variational GP \citep{Hensman:2013} on a binary classification dataset to distinguish background processes from Higgs-Boson particles. Our training data contains one million observations and 28 features and a test set of 100,000 observations with $ J=128 $, $ K=20 $ and $ B=1000 $. Figure~\ref{fig:higgs} shows the comparison of IS-MOE against SVI for this large scale classification task. We see that IS-MOE obtains better predictive log likelihood and AUC scores, while performing slower in comparison to SVI in terms of wall time. Given this fact, it may be possible that the latent generating process for the Higgs-Boson data is non-stationary leading to better performance for IS-MOE over SVI. On the other hand, SVI's apparent superior performance in terms of wall time but not predictive performance could be explained by the fact that there is a local optima that SVI tends to converge to quickly which produces sub-optimal results, whereas IS-MOE must spend more time exploring the posterior space but can find a better result. Nevertheless, the fact that IS-MOE inference procedure can obtain a wall time of about six minutes on a training set of size one million is still quite impressive and it does suggest that IS-MOE can perform well with marginal likelihood approximations to the full model as well. 


\begin{figure}[h]
	\centering
	\includegraphics[width=.9\linewidth]{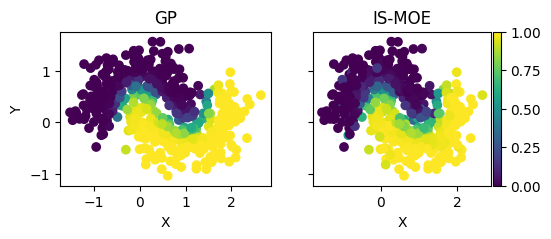}
	\caption{Binary classification task: label probabilities obtained using the full GP and the IS-MOE.}
	\label{fig:classification}
\end{figure}

\begin{table}
	\caption{Test set log likelihood and AUC on three classification datasets.}
	\centering
		\begin{tabular}{lrrrrrr}	
					   &           & Log Likelihood & & & AUC & \\
			\hline

			Data       &   Full GP &   IS-MOE  &   FITC \vline &   Full GP &   IS-MOE  &   FITC \\
			\hline
			Pima       &    -128.79  & -135.09  &      -128.61 \vline &         0.83 &   0.81 &        0.83\\
			Parkinsons &     -17.00 &  -22.76 &       -28.42  \vline&         0.86 &   0.93 &        0.88\\
			WDBC       &     -15.50 &  -12.62 &       -18.01 \vline&         0.83 &   0.91  &        0.81\\
			\hline
		\end{tabular}	
		\label{table:classificationLL}
\end{table}

\section{Summary and future work}\label{sec:conclusion}
While Gaussian processes provide a flexible framework for a wide variety of modeling scenarios, their use has been limited in the ``big data'' regime, since most implementations scale cubically with the number of data points. As discussed in Section~\ref{sec:background}, a number of approximations have been proposed to reduce this cost but these approximations come with notable failure modes. The IS-MOE avoids these pitfalls, using parallelizable importance sampling to explore a mixture of block-diagonal, easily invertible matrices. However, importance sampling can be a rather rudimentary approach to inference. For more sophisticated settings, we may need to resort to particle filters or other sequential Monte Carlo techniques. Of course in such inference methods, the proposal distribution is crucial for performance but difficult to choose in practice. We are interested in investigating the theoretical behavior of proposal distributions in this setting.  

In this paper, we have focused on regression models using a Gaussian mixture model on the covariates, but the scope of the IS-MOE is much broader. For example, we could use alternative distributions over partitions, or embed the IS-MOE within a more complex model--particularly in deep Gaussian process models and non-Gaussian likelihoods. We are also interested in refinements to importance sampling to improve their inferential quality, such as alternative proposal distributions. Another potential avenue for future research is to explore whether we can achieve further speed-ups by using GPU-based computation \citep[for example]{Dai:2014,Gramacy:Niemi:Weiss:2014,Gramacy:Apley:2015}. We leave such explorations for future work.

\section*{Acknowledgments}
Michael Zhang and Sinead Williamson were supported by NSF grant 1447721. Sinead Williamson's contribution was written before her employment at Amazon.
\bibliography{gp}

\end{document}